\documentclass{article}

\usepackage{PRIMEarxiv}

\usepackage[utf8]{inputenc} 
\usepackage[T1]{fontenc}    
\usepackage{hyperref}       
\usepackage{url}            
\usepackage{booktabs}       
\usepackage{amsfonts}       
\usepackage{nicefrac}       
\usepackage{microtype}      
\usepackage{lipsum}
\usepackage{fancyhdr}       
\usepackage{graphicx}       
\graphicspath{{media/}}     

\DeclareUnicodeCharacter{202F}{}

\title{ Assessing Perceived Fairness from Machine Learning Developer's Perspective 
}

\author{
  Anoop Mishra \\
  University of Nebraska \\
  6001 Dodge Street\\
  Omaha, Nebraska \\
  USA - 68182\\
  \texttt{amishra@unomaha.edu} \\
   \And
   Deepak Khazanchi \\
   University of Nebraska \\
   6001 Dodge Street\\
   Omaha, Nebraska \\
   USA - 68182\\
   \texttt{khazanchi@unomaha.edu} \\
}

\begin{document}
\maketitle

\begin{abstract}
 \textbf{Fairness} in machine learning (ML) applications is an important practice for developers in research and industry. In ML applications, unfairness is triggered due to bias in the data, curation process, erroneous assumptions, and implicit bias rendered within the algorithmic development process. As ML applications come into broader use developing fair ML applications is critical. Literature suggests multiple views on how fairness in ML is described from the \textbf{user's perspective} and students as \textbf{future developers}. In particular, ML developers have not been the focus of research relating to perceived fairness. This paper reports on a pilot investigation of \textbf{ML developers’ perception of fairness}. In describing the perception of fairness, the paper performs an exploratory pilot study to assess the \textbf{attributes} of this construct using a systematic focus group of developers.  In the focus group, we asked participants to discuss \textbf{three questions}- 1) What are the characteristics of fairness in ML? 2) What factors influence developer’s belief about the fairness of ML? and 3) What practices and tools are utilized for fairness in ML development? 

The findings of this exploratory work from the focus group show that to assess fairness developers generally focus on the overall ML application design and development, i.e., business-specific requirements, data collection, pre-processing, in-processing, and post-processing. Thus, we conclude that the procedural aspects of \textbf{organizational justice theory} can explain developer’s perception of fairness. The findings of this study can be utilized further to assist development teams in integrating fairness in the ML application development lifecycle. It will also motivate ML developers and organizations to develop best practices for assessing the fairness of ML-based applications.
\end{abstract}


\section{Introduction}
Machine learning (ML) is practiced in the research, industry, and education sectors to make complex decisions and assist humans \cite{mishra2019intelligent, mishra2020interdisciplinary,goodfellow2016deep, phd2022, graca2022}. Machine learning in natural language processing and computer vision are employed in developing decision-making systems and decision-support systems \cite{mishra2020interdisciplinary,goodfellow2016deep}. Measuring fairness and developing fair ML applications has become a widespread practice in research, industry, and academia. However, literature suggest that fairness in ML is a very subjective term and possess many definitions \cite{pessach2022review, mehrabi2021survey, srivastava2019mathematical}. Mehrabi \textit{et al.} 2021 defines fairness in ML as deficiency of favoritism toward an individual or a group based on acquired characteristics \cite{mehrabi2021survey}. Similarly, Pessach \textit{et al.} 2022 describes different notions of fairness like individual and group fairness. The subjective characteristic in fairness is caused due to introduction of bias. An unfair model in ML is triggered mainly due to bias in the data and erroneous assumptions rendered within the algorithmic development process \cite{phd2022}. Literature suggests that bias in data has many shapes and forms. Thus, algorithmic bias and algorithmic fairness are discussed in literature \cite{pessach2022review, mehrabi2021survey, pmlr-v108-jiang20a}. Algorithmic fairness is an area of research applied to mitigate bias and explain fairness in AI systems \cite{mehrabi2021survey,phd2022}. Researchers have multiple views and descriptions for algorithmic fairness, thus lacks a rigid definition. Mehrabi \textit{et al.} 2021 define a large class of biases by representing a feedback loop relationship between data, algorithm, and users \cite{mehrabi2021survey}. They argue that most of the definitions and work on fairness are developed in the West. When these are applied to different problem types, it introduces historical bias, contextual bias, and representation bias \cite{mehrabi2021survey,phd2022}. It may lead to an unfair AI decision-making system. However, few researchers see fairness in ML systems as multi-dimensional aspects like psychology, political science, and economics \cite{wang2020factors, phd2022}. \par

Investigating notions of fairness is important because, if not considered, incorrect outcomes or perceptions may lead to severe societal and business concerns. The literature includes scenarios discussing the severe effect on society: Amazon's AI-based recruiting system reported bias against women in the recruitment process; Apple's credit card approval process biased against women; and Stanford's COVID-19 vaccination algorithms were biased towards a specific group \cite{amazon_2018, covid19, apple, phd2022}.  \par
 
Researchers tried to understand fairness from the sociotechnical domain by conducting studies on the human perception of fairness. The literature describes that different stakeholder has distinct interpretations of fairness concerning the same ML model \cite{lee2017human, phd2022}. These studies are conducted to understand the relationship between human perceptions of fairness and proposed notions of fairness in literature \cite{srivastava2019mathematical, lee2017human, lee2018understanding,lee2017algorithmic, lee2021included, woodruff2018qualitative}. Harrison \textit{et al.} 2022 conducted an empirical user study to investigate the trade-off between competing notions of fairness and human perception of fairness such that ML models can embed these trade-offs to build fair ML applications \cite{harrison2020empirical}. Based on the literature, perceived fairness in ML is described as human perception and interpretation of ML models based on outcomes predicted by the ML model. \par

 Prior studies are conducted to investigate user's perception of fairness. Most of the studies performed a randomized between-subject experiment on Amazon's Mechanical Turk. The validity of research on Amazon's Mechanical Turk has been questioned in the past \cite{thomas2017validity, phd2022}. Kasinidou \textit{et al.} 2021 and Kleanthous \textit{et al.} 2022 investigated student's (as future developers) perception of fairness and justice in algorithmic decision-making in three separate scenarios. However, the potential gap lies within the understanding of the human perception of fairness which has to do with the ML developer's perception of fairness. ML developers are crucial actors to investigate perceived fairness because they are responsible for designing, developing, and evaluating ML models in ML development process. We conducted virtual focus groups to explore and assess characteristics and factors that influence ML developer's perceived fairness. The goal of this research study is to assess the characteristics of perceived fairness from developer's perspective. Thus, we asked three questions from ML developers in the systematic focus groups:
\begin{enumerate}
 \item How would you describe the fairness of ML applications from your (developer's) perspective?
 \item What are the factors that influence your (developer's) belief about the perceived fairness of ML applications?
\item What practices or tools do you utilize to practice fairness?
\end{enumerate}

Inductive thematic analysis and LDA-based topic modeling are utilized to accomplish the research objective of assessing developer's perceived fairness. Section \ref{approach} discusses more about this approach. Our findings assist to understand the relation between actual ML developer's perceptions and proposed notions of fairness in the literature. Researchers explain that notions of fairness are associated with distributed fairness from organization justice theory, which means fairness measures based on outcomes \cite{morse2021ends,grgic2018beyond, pessach2022review}. The present trend in fairness advocates for developing procedural notions of fairness, i.e., procedural fairness that explain fairness based on the process \cite{lee2019procedural,morse2021ends,rueda2022just}. This research study's findings conclude that the developer's perceived fairness relates to procedural fairness of organizational justice theory in the decision-making process. Based on the findings, our research contributions are-
\begin{enumerate}
    \item Developed attributes (themes) of ML developer's perceived fairness,
    \item Investigated the association between ML developer's perceived fairness with procedural fairness,
    \item Proposed definition of perceived fairness from ML developer's perspective
\end{enumerate}

This research study's findings will help researchers and motivate ML developers and practitioners to understand perceived fairness from a multi-dimensional perspective. This research will also help organizations understand perceived fairness and provide insight into how fairness is addressed when interpreting ML applications. The latter section includes the related work discussing prior research on fairness, practice and tools in fairness, and human perception (users and students) of fairness. The related sections also discuss the gaps in existing methods. Section \ref{Methods} shows the approach of inductive thematic analysis. Further sections include findings and discussions, association of procedural fairness and perceived fairness, definition of perceived fairness, and conclusions.


\section{Background}
\subsection{Algorithmic fairness in decision-making}
In artificial intelligence (AI), machine learning (ML) is described as data-driven process that learns with experiencing the data without explicitly programmed \cite{goodfellow2016deep, mishra2019intelligent}. The advancement in ML approaches allows models to assist humans in decision-making tasks. Useful representations are learned from data that evaluates the ML model's performance. However, if the data consist biases then models from data-driven process may inherits the bias. Since, ML models like neural networks are black box that causes intermediate process to be opaque, it becomes difficult to assess whether the decisions are justified or biased \cite{phd2022,mehrabi2021survey}. Prior research indicates critical concern on data collection methods, because flawed data include bias which can result in unfair decisions \cite{mehrabi2021survey,google, pessach2022review}. Thus, it is essential to understand the definition of a fairness in ML. Literature suggests that ML model is unfair if it produces unfavorable treatment to people based on specific demography \cite{dwork2012fairness,kusner2017counterfactual,beutel2019putting,phd2022}.
Fairness in ML is a popular and multi-dimensional concept that depends on cultures, objectives, contexts and problem definition \cite{wang2020factors,lee2017human,biran2017explanation,doshi2017towards}. Black box nature of ML models that lacks explainability can lead to harmful consequences. Techniques utilizing computational and mathematical frameworks are utilized to practice interpretability, bias identification, that improves algorithmic fairness, such as IBM's AI360, Tensorflow constrained optimization framework, fairlearn, etc. \cite{del2020review,vasudevan2020lift, bellamy2018ai, deng2022exploring, phd2022}.
Deng \textit{et al.} 2022 empirically explored the practice of ML toolkits from ML practitioners. They concluded that practitioners need toolkits such that they can contextualize, collaborate and communicate in explainability with non-technical peers for ML fairness \cite{deng2022exploring}.


\subsection{Human perception of fairness} \label{litper} 
Prior research advocates that for developing a fair decision-making system consult with subject matters, ethical checks, planning, and human checks must be considered \cite{covid19,google,mishra2020interdisciplinary, phd2022}. Lee \textit{et al.} 2017 finds that different stakeholders perceive distinct interpretations of fairness with same ML model \cite{lee2017human}. For example, an online experiment conducted by Wang \textit{et al.} focusing AI/ML fairness conclude that if algorithmic outcomes are inclined towards an individual, then it was rated as fair by the users \cite{wang2020factors, phd2022}. Researchers made efforts to understand fairness from the sociotechnical domain by conducting studies to investigate human’s perception of fairness. These studies are organized to understand the relationship between human’s perception of fairness and proposed notions of fairness in literature \cite{srivastava2019mathematical, lee2017human, lee2018understanding,lee2017algorithmic, lee2021included, woodruff2018qualitative}. Woodruff \textit{et al.} 2018 explored the perceived fairness of users from a marginal population, they found that ML fairness cooperates with user's trust \cite{woodruff2018qualitative}. Berkel \textit{et al.} 2021 performed an online crowdsourcing study to investigate how information presentation influences human’s perceived fairness \cite{van2021effect}. Srivastava \textit{et al.} 2019 conducted user's study to investigate the relationship between mathematical notions and human perceived fairness \cite{srivastava2019mathematical}. Lee \textit{et al.} 2021 performed user’s experiments focusing on people’s (Black American) perception of fairness in AI healthcare for skin cancer diagnosis. This research seeks to examine individual-level differences targeting trustworthiness in human decision-making in contrast to AI-based algorithmic decision-making \cite{lee2021included}.

\subsection{Fairness from organization justice theory} \label{justicetheory}
Piero \textit{et al.} 2014 explains that fairness is concerned with social norms and governing rules \cite{Peiro2014, phd2022}. They discuss four forms of fairness considering the fourfold model of justice theory: distributive justice, procedural justice, interpersonal justice, and informational justice. They claim that perceived fairness is highly correlated with psychological well-being and distress. Algorithmic fairness is broadly explored with distributed (justice) fairness. Few attempts are made to explore procedural fairness \cite{lee2019procedural,morse2021ends}. Distributive Fairness is defined as perceived fairness is the process for distribution of rewards across group members \cite{cohen1987distributive, mcfarlin1992distributive, greenberg1987taxonomy, grgic2018human, phd2022}. Whereas procedural justice is the perceived fairness of rules and decision processes used to determine outcomes \cite{mcfarlin1992distributive, lee2019procedural, Leventhal1980, greenberg1987taxonomy, grgic2018beyond}. 
\subsubsection{Procedural Fairness}
Few studies explain the advantage of procedural fairness over distributed fairness. Morse \textit{et al.} 2021 and Rueda 2022 suggest that procedural fairness in ML augments explainability and transparency in the ML model \cite{rueda2022just,morse2021ends}. Biran \textit{et al.} 2017 claims that bias and fairness are highly related to the interpretation and explainability \cite{biran2017explanation, phd2022}. In machine learning, the explanation is complex due to its black-box nature. Doshi-Velez \textit{et al.} 2017 explain the relation between interpretability with reliability and fairness by discussing real-world scenarios \cite{doshi2017towards, phd2022}. The definition of fairness, explainability, and interpretability is motivated by multiple theories of lens, including psychology, philosophy, cognitive science, and ethics\cite{doshi2017towards,mishra2019intelligent,mishra2020interdisciplinary, phd2022}. 
However, in this study, we are targeting organization justice theory. We identified components of procedural fairness from the literature to accomplish our objective. Lee \textit{et al.} 2019 describes procedural fairness using transparency, control, and principle \cite{lee2019procedural}. Rueda 2022 explains procedural fairness as avoidance of bias, accountability, and transparency \cite{rueda2022just}. Morse \textit{et al.} 2021 discuss the components of procedural fairness from Leventhal 1980 as bias impression, consistency, representativeness, correctability, accuracy, and ethicality \cite{Leventhal1980, morse2021ends}. These components are utilized and discussed in detail in section \ref{definition}.
\par
Most of the studies discussed in section \ref{litper} include participants as users and students. Few studies utilized Amazon Mechanical Turk for user's study. However, the validity of research on Amazon's Mechanical Turk has been questioned in the past \cite{thomas2017validity, phd2022}. There are no studies that explored ML developer's perception of fairness. Literature suggests that consulting with subject matter experts, ethical checks, planning, and human checks must be considered for developing a fair decision-making model \cite{google,mishra2020interdisciplinary, phd2022}. Studying perceived fairness from organization justice theory as a theory of lens will help define and describe attributes of perceived fairness and develop a conceptualization of the factors that influence the beliefs of ML developers. Thus, it is essential to understand the characteristics of perceived fairness from ML developer’s perspective.


\section{Methods} \label{methods}

\subsection{Virtual Focus Group}
A focus group in research is a group discussion of people with similar characteristics where they share experiences and discuss in order to generate data \cite{kitzinger1995qualitative, khazanchi2006patterns}. Focus group discussions are utilized as a qualitative approach to assess in-depth understanding of social issues \cite{o2018use, khazanchi2006patterns}. In this research study, focus groups are used to explore ML developer’s perceptions of fairness. The participants targeted for the focus groups are ML developers, data scientists, and ML engineers from the industry who participate in the design and development of ML applications. 
 
\subsubsection{Focus group participants}
Motivated by literature and to explore ML developer’s attributes of perceived fairness, we conducted three virtual focus groups on Zoom. Anonymity is ensured for all companies including participants, companies, and intermediate representatives. Table \ref{table 2} shows the participation details for each company.
. 

\begin{table*}[htp]
\centering
\caption{Participants in the focus groups}
\label{table 2}
\begin{center}
\begin{tabular}{|p{0.5in}|p{1.7in}|p{1.2in}|p{1.2in}|}
\hline
\textbf{Company ID} & \textbf{Type of Company} & \textbf{Participants committed to participate} & \textbf{Participants who actually participated}\\
\hline
1
&
Life insurance, finance, medicare
& 
5
&
3
\\
\hline
2
&
Railroad
& 
11
&
3
\\
\hline
3
&
Transportation and logistics 
& 
Unknown
&
3
\\
\hline
\end{tabular}
\end{center}
\end{table*}

Three companies located in United States participated in the virtual focus groups. The company size is more-than-1000-employees for all three companies. The number of participants who actually participated are less with participants committed to participate. In total, nine participants from three distinct companies participated in three virtual focus groups. The diversity of the companies are shown in table \ref{table 2}. The job-role portfolio of the participants includes data scientists, senior data scientist, and software developers. Out of 9 participants, 3 participants are females and 6 are male. Out of all, $75\%$ were extremely familiar with ML concepts including professional ML experience of 3 to 5 years, and took the college-level course of ML. $25\%$ participants are experts and have more than 5 years of professional experience in ML development. 
\subsubsection{Study Design}
The research has Institutional Review Board (IRB) approval for conducting the focus groups. The virtual focus group was designed to develop an understanding of perceived fairness and it’s attribute among developers. This research study utilized MIRO as a brainstorming tool and Zoom to conduct the focus groups. MIRO is an online visual platform where teams can connect, collaborate, create, and brainstorm together (see \url{https://miro.com/}). All sessions are conducted synchronously. Each company along with its participants has one focus group. Each focus group is given a 75-minutes window to participate. Participants from each company were provided with their own unique session ID on MIRO. Participants were notified by email along with MIRO and Zoom web links in advance of the session, so they could plan for their participation. Reminder e-mails were sent during the session to encourage more participation. 
Each company is assigned one focus group. After logging in at Zoom, each company with participants receives an introductory session about the focus group agenda. Each participant is asked to first fill out a questionnaire for demographic details. The second step on the agenda is brainstorming questions, and the participant is asked to enter as many ideas as possible on MIRO during the 75-minutes window, as well as comment and discuss fellow participant’s ideas. The participants recorded their ideas and discussion in the form of notes at MIRO. These notes include phrases, short sentences, and long sentences. The final step on the agenda is a closing discussion based on the brainstorming session. Table \ref{agenda} shows the agenda and instructions for each focus group. The table \ref{agenda} was repeated for each company as seen in table \ref{table 2}.\\


\begin{table}[htp]
\centering
\caption{Focus group agenda and instructions for participants}
\label{agenda}
\begin{center}
\begin{tabular}{|p{1.5in}|p{3.7in}|}
\hline
\textbf{Activity} & \textbf{Instructions}\\
\hline
Introductory session
& 
Welcoming the participants; Making participants aware of Institutional Review Board (IRB) approval for conducting the focus group; Introduction to the focus group with agendas; Introduction of the topic and research objective; Tools introduction and demo; Request to fill out demographic questionnaire; \\
\hline
Beginning Questionnaire
&
Please answer all questions to the best of your knowledge. All questions are voluntary. After you have answered all questions, "Notify" the host, and you will be taken back to the agenda. When you are back at the agenda, go to Brainstorming Question 1 and you can start discussing and entering ideas.\\
\hline
Brainstorming Question 1
&
How would you describe the fairness of ML applications from your perspective? Alternatively, what are the characteristics of the perceived fairness of ML applications? Think broadly to include individual behaviors and processes based on your past experiences like projects and team meetings. Enter 5 to 10 separate ideas. Comment on ideas other people have entered and/or enter more of your own ideas. Feel free to expand on other people’s ideas.\\
\hline
Brainstorming Question 2
&
What are the factors that influence your belief about the perceived fairness of ML applications? Think broadly to include individual behaviors and processes based on your past experiences like projects and team meetings. Enter 5 to 10 separate ideas. Comment on ideas other people have entered and/or enter more of your own ideas. Feel free to expand on other people’s ideas.\\
\hline
Brainstorming Question 3
&
What practices or tools do you utilize to mitigate bias and practice fairness in ML application development? Think broadly to include individual behaviors, processes, technologies, and tools based on your past experiences like projects and team meetings. Enter 5 to 10 separate ideas. Comment on ideas other people have entered and/or enter more of your own ideas. Feel free to expand on other people’s ideas. \\
\hline
Closing Discussion
&
Summarizing all participant's ideas and requesting their agreement for closing the agenda.\\
\hline
\end{tabular}
\end{center}
\end{table}

\subsection{Developing attributes/themes from focus group's data} \label{approach}
 \subsubsection{Thematic analysis}
The focus group data collected from the participant's brainstorming are qualitative in nature. Thus, the data are further analyzed utilizing thematic analysis. Thematic analysis is broadly used for the analysis of qualitative data for recognizing different patterns and allows researchers to formulate rich, detailed, and transparent meanings \cite{braun2006using}. Braun et al. \cite{braun2006using} explain that thematic analysis uses familiarization, code formulation, generation of themes, themes review, defining and naming themes, and report formation. In this research study, the thematic analysis is used to identify emerging themes to formulate and assess the perceived fairness of ML developers. 

\subsubsection{Topic Modeling}
Topic modeling is a statistical modeling approach for discovering abstract “topics” in a collection of documents like newspapers and digital corpus \cite{li2021bibliometric}. Latent Dirichlet Allocation (LDA) approach is used for conducting the topic modeling on focus group data. LDA is a popular topic modeling technique to extract topics from the collection of documents \cite{blei2003latent}. The details of LDA-based topic modeling can be found at \cite{blei2003latent}.

An inductive approach is used to derive the themes by coding qualitative data into clusters of similar entities and conceptual categories. The theme derivation is done by integrating thematic analysis and LDA-based topic modeling approach. The themes helped to formulate the theoretical explanation and definition of ML developer's perceived fairness in machine learning.

\section{Findings and Discussions}

\subsection{Results and findings} \label{results}
In this section, we present the findings from the focus groups data collected from ML developers. The focus groups suggest that participant's ideas and discussions are influenced by their personal experience, knowledge base, and practice gained through developing ML applications. Inductive approach using thematic analysis and topic modeling using LDA assisted to derive themes from the developer's discussion on focus groups. Table \ref{themes} describes the derived themes including their attributes which describe the sub-themes derived from the focus group data. The supportive evidence from focus groups in table \ref{themes} shows the transcripts from focus group discussions. These themes are bias mitigation, data, model design, model validity, business rules, and users interaction, which describe the attributes of the developer's perceived fairness. All the themes, based on ML developer's discussion are described below.

\begin{table*}[htp]
\centering
\caption{Themes derived from thematic analysis and LDA topic modeling with transcripts as supporting evidence}
\label{themes}
\begin{center}
\begin{tabular}{|p{1in}|p{2in}|p{2.5in}|}
\hline
\textbf{Themes} & \textbf{Attributes} & \textbf{Supporting evidence from focus groups}\\
\hline
Bias Mitigation
&
Historical bias, asymmetric bias, selection bias, unintended bias, human bias, implicit bias, outliers
& 
1) "Data wrangling and derivations are not done in such a way as to be cherry-picking data or unduly biasing the results based on human desire". 
\\
\hline
Data
&
Demographics, population, data source, sampling, protected category, balanced/unbalanced data, representation, data review, data diversity, data collection, anonymity, features  
& 
1) "if the data is coming into the model is skewed towards a certain group, the results will reflect it". \newline
2) "I know there is a belief that machine learning models are biased. To me if the coming data is unbalanced, but the model isn't doing anything to skew the output results, it is fair".

\\
\hline
Model Design
&
Algorithmic selection, Adaptability, Blank Slate, Model structure, hyperparameters, auto/manual design, active design 
& 
1) "active design changes should be considered for the sake of fairness, as unfair". \newline
2) "choose the appropriate ML algorithm for the data such that if data is balanced vs unbalanced Data in any stratification factors, use a ML Model appropriate for that design". 
\\
\hline
Model validity
&
Residual analysis, performance metrics, human feedback, explainability, risk assumption, output measures, human choices, boundary conditions  
& 
"Microsoft Azure studio is utilized as a tool for practicing fairness, as the developers stated that for outcome analysis it has a feature to control fairness by ensuring the accuracy/recall is similar across the protected groups."
\\
\hline
Business Rules
&
Project requirements, business constraints, user’s feedback, requirements, construct, user’s usability, use-case analysis, goal-specific selection, target objective, explainability, ethics, privacy
& 
"training data used to construct the model must precisely represent the requirements of the business product."
\\
\hline
User interaction
&
user’s feedback, user’s usability,  explainability to users, case dependent
& 

"one more concern with data insufficiency is when populating the data with median or average for the crucial features, results are not acceptable by the users. So, data MUST be collected appropriately by the application. In Freight Acquisition model (FAM) which is currently deprecated due to data inconsistencies, users believed that model would be giving exact yes/no to call a customer. It took some time to explain the process".
\\
\hline

\end{tabular}
\end{center}
\end{table*}

\subsubsection{Bias mitigation}:  An unfair model in ML is affected due to bias in the data \cite{pessach2022review, mehrabi2021survey, pmlr-v108-jiang20a}. The findings implies that developers are concerned with training the ML model which is a true representative of the population. Key bias form that involved in the discussion are historical, unintended (bias due to location), implicit bias, and human annotated bias. The bias mitigation techniques used by the developers include hold-out sampling, rigorous residual analyses, chi-square tests on residuals (to ensure no statistically significant patterns exist in residual distributions), implicit bias by the protected classification, and differences in the model predictions across groups.

\subsubsection{Data}: This theme explain data and its properties that highly depend on how data is collected. The developer's perception suggests that data collection (sampling) is important to target true representatives of the population. ML developers discussed how data is stored, processed, and transmitted in ML development process. Data representation in input data and project requirements is important to build an ML model.  Feature engineering and data wrangling techniques are used widely to understand the data. They discuss that feature engineering helps them to analyze how certain variables are being weighted based on the historical understanding of the data/problem. Key data representation practices discussed are data transformation, cross-validation, data wrangling, and dimensionality reduction. 

\subsubsection{Model design}: Model design reveals the model development process incorporating algorithmic selection, parameters selection, and training of the ML model. ML Developer's discussed that developers must be a "blank slate" while performing the quantitative evaluation of the model with proper metrics such that no bias through the developer's action can be logged.

\subsubsection{Model validity}: This theme illustrates whether the ML model accomplishes its intended business objective or not. The developer's discussion enlightens that developing explainable models is their key aspect of practicing fairness. Key practices discussed are risk assumptions from use-case testing, evaluating fairness from the true objectives of the scenario, peer review of code from fellow developers, marginal analysis, analyzing true predictions based on demographics, human-in-loop, and evaluating model performance with multiple metrics over time. 

\subsubsection{Business rules}: Business rules are framing the ML problem by defining constraints, rules, ethics, privacy, and stakeholder's goals.  The developer's discussion explains that the true objectives and goals of the ML must be set such that evaluations can be done within the boundaries of business rules, not on human choices. They explained that the development should not include features that violate privacy. 

\subsubsection{Users Interaction}: The findings suggest that developers are users oriented. The developers discuss that the business directly deals with people. Thus, a fair ML  model should be beneficial to end-users, not biased towards certain groups, and being explainable to users. One of the important aspects of their perceived fairness was explaining the ML flow process to users and domain experts. They further explained that user’s feedback is recorded and then integrated with the ML process such that ML applications are built with the appropriate context. 

\textbf{Developers discussion on \textit{"fairness"}}: Based on the findings of our study  and the discussion above, we conclude that the developer’s perceived fairness comprises the complete ML process including privacy, ethics, the intention of ML development, business constraints and goals, explainability to users, and user’s usability. Interestingly, one of the developers claims that fairness in machine learning is a subjective term and evaluation of ML models must include ML-pipeline process.

\subsection{Defining Perceived Fairness} \label{definition}

In section \ref{justicetheory}, we reviewed the components of procedural fairness from organization justice theory, discovered from literature. Lee \textit{et al.} 2019 describes procedural fairness using transparency, control, and principle \cite{lee2019procedural}. As per Lee \textit{et al.} 2019, transparency is the rules of the decision-maker that are perceived as fair and warranted including an explanation of decision outcomes and information representativeness. Control is described as the degree of control over the decision that individuals receive, and principle is defined as demonstrations of consistency, competency, benevolence, and voice. Rueda 2022 explains procedural fairness as avoidance of bias, accountability, and transparency in medical scenarios \cite{rueda2022just}. Rueda 2022 defines transparency as the procedure that explains ML algorithms working and processing that lead to the outcome. Accountability is also related to the robustness of the model and avoidance of bias describes not including attributes that can cause unfavorable decisions \cite{rueda2022just}. Morse \textit{et al.} 2021 discuss the components of procedural fairness proposed by Leventhal 1980 as bias impression, consistency, representativeness, correctability, accuracy, and ethicality \cite{Leventhal1980, morse2021ends}. Consistency defines the uniformity of decision procedures across people and time, accuracy is the measure of validity and high-quality information, ethicality describes practicing moral standards and values, representativeness describes proper population representation, bias suppression subjects to prevent favoritism by the decision maker, and lastly, correctability are approaches to correct flawed decisions \cite{Leventhal1980,morse2021ends}. \par
Table \ref{Procedural} shows an association of themes describing developer's perceived fairness with the components of procedural fairness proposed by Lee \textit{et al.} 2019, Rueda 2022, and Leventhal 1980. These associations are proposed by the union of the procedural fairness component's description from the literature discussed and the ML developer's discussion in the focus groups. For example, the theme "Data" describes the properties and characteristics of data in ML development process, as discussed in Sections \ref{results}. This theme aligns with Lee's \textit{et al.} 2019 transparency and control because of information representativeness and its impact on decisions in the data-driven process. Rueda's 2022 avoidance of bias for fair decision-making, and Leventhal's 1980 consistency and representativeness for uniform decision-making across people and time.
Thus, we conclude that the association in table \ref{Procedural} illustrates that procedural aspects of organizational justice theory, i.e., procedural fairness can explain the developer’s perception of ML fairness.

\textbf{Definition of \textit{perceived fairness}}-  In section \ref{results}, we discussed the characteristics of perceived fairness from the developer's perspective. Based on the findings of this study, a developer's perception of fairness relates to aspects of data, user characteristics, understanding of ML model design and validity, and understanding of business rules that impacts their behaviors and ability to build ML systems that are free from bias. This implies that the ML systems are designed and built to be fair in processes, transparent in actions (explainable), have the opportunity for multiple voices to be integrated into their development, and are impartial to all users in their outcomes.

\begin{table*}[htp]
\centering
\caption{Relationship between themes of fairness and procedural (justice) fairness components from literature}
\label{Procedural}
\begin{center}
\begin{tabular}{|p{1in}|p{1.3in}|p{1.5in}|p{1.7in}|}
\hline
\textbf{Themes} & \textbf{Lee et al. 2019 } & \textbf{Rueda 2022 } & \textbf{Leventhal 1980; Morse et al 2020 }\\
\hline
Bias Mitigation
&
Control  
&
Avoidance of bias, Accountability 
&
Bias suppression 
\\
\hline
Data
&
Control, Transparency
&
Avoidance of bias 
&
Consistency, Representiveness 
\\
\hline
Model Design  
&
Transparency 
&
Transparency 
&
Correctability 
\\
\hline
Model Validity 
&
Transparency, Control, Principle 
&
Transparency, Accountability 
&
Accuracy, Correctability 
\\
\hline
Business Rules
&
Transparency, 
&
Avoidance of bias, Accountability 
&
Ethicality 
\\
\hline
Users Interaction
&
Transparency, Control, Principle 
&
Accountability 
&
Ethicality 
\\
\hline
\end{tabular}
\end{center}
\end{table*}

\section{Implications and Future Works}
\subsection{Implications} \label{Implications}
\begin{table*}[htp]
\centering
\caption{Explaining relationships between attributes of perceived fairness of developers and ML development process}
\label{MLthemes}
\begin{center}
\begin{tabular}{|p{1.5in}|p{2in}|}
\hline
\textbf{Themes} &  \textbf{Relationship with ML process}\\
\hline
Bias Mitigation  
&
Define and plan, pre-processing
\\
\hline
Data 
&
Define and plan, pre-processing  
\\
\hline
Model Design
&
In-processing 
\\
\hline
Model Validity  
&
Post-processing  
\\
\hline
Business Rules
&
Define and plan
\\
\hline
Users Interaction
&
 Define and plan, post-processing 
\\
\hline
\end{tabular}
\end{center}
\end{table*}

In this research study, we acknowledge the relationship between themes mentioned in table \ref{themes} with the ML development process utilized to develop ML models and applications. The relationship shown in table \ref{MLthemes} is validated by the definitions discussed in the literature \cite{AWS}. In literature, the ML process is divided broadly into define-and-plan, pre-processing, in-processing, and post-processing \cite{goodfellow2016deep, janiesch2021machine, ge2017data}. All these ML processes have distinct objectives for any framed ML problem \cite{AWS}. The findings will help and motivate ML practitioners, researchers, and organizations to develop and further explore the research for formulating fairness in ML. This research study will also help researchers to understand the perceived fairness in ML in a realistic setting and provide insights into how perceived fairness is addressed while evaluating ML applications.  It will also motivate other ML developers and organizations to develop and practice ethical and fair ML decision-making models to benefit society and businesses.
 

\subsection{Future Works}
This is a \textbf{work-in-progress} article. As observed, there are only 9 participants, and a focus group is organized just to perform a pilot study of the existing research. However, the results advocate some promising future directions including performing a large survey study.

\section{Conclusion}
In this \textbf{pilot research study}, we explored the ML developer's perception of fairness using pilot investigation through focus groups. Three companies along with nine ML developers participated in focus groups. An inductive approach, integrating thematic analysis and LDA-topic modeling is utilized to derive themes that describe attributes of ML developers' perceived fairness. The findings of the study conclude two major arguments- 1) developer's perceived fairness generally focuses on the overall ML application design and development, i.e., business-specific requirements, pre-processing, in-processing, and post-processing. 2)  the procedural aspects of organizational justice theory can explain developer’s perception of fairness. Finally, we proposed the definition of perceived fairness from ML developer’s perspective.


\section{Acknowledgement}
This pilot research study acknowledges and thanks all the companies and their participants along with their managers whose participation made this study possible. This article acknowledges the GRACA 2021 grant under the title \textit{Perceived Fairness from Developer’s Perspective in Artificial Intelligent Systems}. The writings of this research paper were included in the GRACA grant application and oral presentation at the research fair presentation of the Office of Research and Creativity Activity under the title \textit{Perceived Fairness from Developer’s Perspective in Artificial Intelligent Systems} at the University of Nebraska at Omaha (UNO) in Spring 2022. The virtual focus group for this research study has been approved by the Institutional Review Board (IRB 263-21-EX).

\bibliographystyle{unsrt}  
\bibliography{references}

\end{document}